\documentclass[letterpaper]{article} 
\usepackage{aaai24}
\usepackage{times}  
\usepackage{helvet}  
\usepackage{courier}  
\usepackage[hyphens]{url}  
\usepackage{graphicx} 
\usepackage{natbib}  
\usepackage{caption} 
\usepackage{algorithm}
\usepackage{amsmath}
\usepackage{graphicx}
\usepackage{url}
\usepackage{xcolor}
\usepackage{comment}
\usepackage{caption}
\usepackage{algpseudocode}
\usepackage{adjustbox}
\usepackage{multirow}
\usepackage{booktabs}
\usepackage{bibentry}
\usepackage{newfloat}
\usepackage{listings}
\DeclareCaptionStyle{ruled}{labelfont=normalfont,labelsep=colon,strut=off} 
\floatstyle{ruled}
\newfloat{listing}{tb}{lst}{}
\floatname{listing}{Listing}
\title{Towards Reliable AI Model Deployments:\\Multiple Input Mixup for Out-of-Distribution Detection}
\author {
    Dasol Choi$^{*1}$ \;
    \stepcounter{footnote}Dongbin Na$^{*2}$\thanks{Correspondence to dongbinna@postech.ac.kr}
}
\affiliations{
    $^1$Kyunghee University \;
    $^1$MODULABS \;
    $^2$Pohang University of Science and Technology (POSTECH)
}

\begin{document}

\maketitle
\def\thefootnote{*}\footnotetext{These authors contributed equally to this work.}
\vspace{-0.1cm}
\begin{abstract}
\vspace{-0.2cm}
Recent remarkable success in the deep-learning industries has unprecedentedly increased the need for reliable model deployment.
For example, the model should alert the user if the produced model outputs might not be reliable.
Previous studies have proposed various methods to solve the Out-of-Distribution (OOD) detection problem, however, they generally require a burden of resources.
In this work, we propose a novel and simple method, Multiple Input Mixup (MIM).
Our method can help improve the OOD detection performance with only single epoch fine-tuning.
Our method does not require training the model from scratch and can be attached to the classifier simply.
Despite its simplicity, our MIM shows competitive performance.
Our method can be suitable for various environments because our method only utilizes the In-Distribution (ID) samples to generate the synthesized OOD data.
With extensive experiments with CIFAR10 and CIFAR100 benchmarks that have been largely adopted in out-of-distribution detection fields, we have demonstrated our MIM shows comprehensively superior performance compared to the SOTA method.
Especially, our method does not need additional computation on the feature vectors compared to the previous studies.
All source codes are publicly available at \textcolor{blue}{\url{https://github.com/ndb796/MultipleInputMixup}}.
\vspace{-0.2cm}
\end{abstract}
\section{Introduction}
\vspace{-0.1cm}
As deep-learning models have achieved remarkable success across various domains, the reliability of AI has been a greatly important issue.
However, modern deep neural networks often show high confidence in their predictions, even when faced with Out-of-Distribution (OOD) data.
This is a critical problem, as models should ideally alert users to potentially unreliable or inaccurate predictions.
For example, consider an autonomous vehicle encountering objects that were not included in its training dataset. In such scenarios, the ability to accurately distinguish between In-Distribution (ID) and Out-of-Distribution (OOD) data becomes crucial.

While various OOD detection methods have been proposed recently, they generally focus on the post-hoc approach~\cite{MSP, mahalanobis, liang2017enhancing} utilizing output vectors or features extracted from the \textit{fixed pre-trained classifier}.
These methods do not alter the weight parameters of the model, thus preserving the original classification performance.
However, they usually require an additional proxy OOD dataset to fit the method to the dataset, which may not be practical in real-world scenarios.
In this work, we propose a realistic and straightforward approach: Multiple Input Mixup (MIM), which only requires a single additional training epoch post-training. 
This method provides a practical solution with minimal computational burden, especially compared to the extensive training iterations common in computer vision domains.

Previous studies~\cite{kirby,gan} have shown that the marginal features can be utilized as artificial OOD samples.
However, these methods typically involve additional neural networks or significant training overhead, which might be not feasible.
To overcome these limitations, we have explored various data augmentation methods for out-of-distribution (OOD) detection.
During this procedure, we have observed that the general input mixup provides useful information for data augmentation~\cite{mixup}. Notably, mixing multiple samples, particularly more than five images can produce semantic features significantly different from the original image manifold.
Motivated by this observation, we propose a simple, yet effective method, Multiple Input Mixup (MIM), that mixes various input samples to generate synthetic OOD samples.
With just a single epoch of training on these artificially created OOD samples using a pre-trained classifier, MIM demonstrates superior OOD detection performance across various benchmarks.
Our contributions are as follows:
\vspace{-0.1cm}
\begin{itemize}
    \item In this work, we propose a simple yet effective method, MIM, that improves the OOD detection performance by mixing multiple ID samples.
    \item Our method requires only an additional single epoch for the pre-trained model and has minimal impact on the original classification performance.
    \item Despite its simplicity, our method shows competitive OOD detection performance compared to state-of-the-art methods.
\end{itemize}
\vspace{-0.3cm}

\section{Related Work}
\vspace{-0.1cm}
Previous studies in the OOD detection research fields can generally be divided into two aspects: (1) post-hoc methods and (2) OOD data utilization methods.
First, the post-hoc methods~\cite{odin,entropy,energy,KLmatchingLogits,vim,mahalanobis,MaxLigit,DICE,RMD} do not require \textit{additional training} for the trained model.
These methods typically utilize feature vectors or outputs extracted 
from the pre-trained model without modifying the model's parameters.
However, some post-hoc methods~\cite{mahalanobis} involve tuning the OOD detection hyper-parameter with a small subset of oracle OOD datasets, which might not be feasible in some real-world deployment settings.
Moreover, certain post-hoc methods also demand additional steps for feature manipulation~\cite{KLmatchingLogits,mahalanobis}.
Our method does not require an extra explicit oracle OOD dataset, providing a more practical solution.

Secondly, various OOD data utilization methods have been proposed for OOD detection.
They usually require the use of explicit OOD datasets for training.
While they demonstrate high OOD detection performance, accessing real-world OOD datasets is fundamentally challenging.
One study, showing competitive performance with Outlier Exposure (OE)~\cite{OE}, requires additional steps for feature marginalization, which require a burdensome extra data synthesizing process.
Furthermore, various OOD data utilization methods typically need intervention during the training time. 
For instance, Outlier Exposure (OE) necessitates the use of both ID data and explicit OOD data throughout the training period.
In contrast, our method only requires a single additional epoch for the pre-trained classifier.
Our method achieves superior OOD detection performance while maintaining the original model classification performance.

\section{Proposed Methods}
\vspace{-0.1cm}
\subsection{Multiple Input Mixup}

The previously proposed strategies utilizing the Mixup approaches~\cite{mixup, cutmix, manifoldmixup} primarily employ the technique of mixing two data samples to generate realistic images that resemble real-world examples. These methods mainly focus on improving the test accuracy of ID datasets and have not been widely used to enhance OOD performance.

In this work, we propose a novel method, multiple input mixup.
Unlike traditional Mixup methods that mix two data samples, our approach involves combining a larger number of samples.
Given $n$ data samples $\{x_1, \ldots, x_n\}$ as a training dataset, we simply mix $k$ samples, where $k$ is typically much greater than 5, often around 10.
Ours generates data samples that contain marginal information, which refers to features that are less discriminative or less indicative of specific class characteristics.
We define these images containing marginal feature information as OOD data and employ them in our training process.
Specifically, class-discriminative features, which are more closely related to specific classes, are considered ID data, while marginal information images generated using our MIM are categorized as OOD data. 

Remarkably, when we apply this approach to benchmarks such as CIFAR-10 and CIFAR-100 where ten or more samples are mixed, we observe significant improvements in out-of-distribution (OOD) performance.
To the best of our knowledge, our paper is the first to demonstrate the utility of mixing more than ten data, showing its potential benefits.
\vspace{-0.1cm}
\begin{algorithm}
\caption{Multiple Input Mixup (MIM) with Augmentation Utilizing a Single Training Epoch}
\begin{algorithmic}[1]
\State \textbf{Input:} CIFAR-10 dataset $\mathcal{D}$, the classification model $M$, learning rate $\alpha = 0.0001$, momentum $\mu = 0.9$, the size of mix-up $m$
\State \textbf{Output:} Fine-tuned model $M$, OOD detection performance metrics
\State \textbf{Initialize the model:} $M \gets \text{pre-trained on CIFAR-10}$
\State \textbf{Load datasets:} 
\State \;\;\;$\mathcal{D}_{ID} \gets \text{CIFAR-10}$, $\mathcal{D}_{OOD} \gets \text{external datasets}$
\State \textbf{Define detector and optimizer:} 
\State \;\;\;$\mathcal{S} \gets \text{MaxSoftmax}(\cdot) \text{ based on } M$
\State \;\;\;$\mathcal{O} \gets \text{SGD with } \alpha, \mu$
\State \textbf{Set batch size:} $B$
\State \textbf{Define augmentation transformation:} $\mathcal{T}$

\For{\textbf{each batch} $(x_{\text{ID}}, y_{\text{ID}})$ \textbf{from} $\mathcal{D}_{ID}$}
    \State \textbf{Compute outputs:} $\mathbf{o}_{\text{ID}} \gets M(x_{\text{ID}})$
    \State \textbf{Calculate original loss:} $\mathcal{L}_{\text{ID}} \gets \text{CE}(\mathbf{o}_{\text{ID}}, y_{\text{ID}})$
    \State \textbf{Apply mix-up:} $x_{\text{mix}} \gets \text{MixUp}(x_{\text{ID}}, m)$
    \State \textbf{Apply augmentation:} $\mathbf{x}_{\text{aug}} \gets \mathcal{T}(x_{\text{mix}})$
    \State \textbf{Compute augmented outputs:} $\mathbf{o}_{\text{OOD}} \gets M(x_{\text{aug}})$
    \State \textbf{Calculate mix-up loss:} 
    \State \;\;\;$\mathcal{L}_{\text{OOD}} \gets \text{CE}(\mathbf{o}_{\text{OOD}}, \mathcal{P}_{Uniform})$
    \State \textbf{Update model:} $\mathcal{O}.\text{step() with } \mathcal{L}_{\text{ID}} +\mathcal{L}_{\text{OOD}}$
\EndFor
\State \textbf{Evaluate the fine-tuned model:} 
\State \;\;\;$\text{Evaluate the model utilizing $\mathcal{S}$ on } \mathcal{D}_{\text{OOD}} \text{ and } \mathcal{D}_{\text{ID}}$
\end{algorithmic}
\end{algorithm}
\vspace{-0.1cm}

\subsection{Simple Fine-tuning}
\vspace{-0.1cm}
Our method leverages fine-tuning~\cite{finetuning} the pre-trained model $M$ for just a single epoch, which requires nearly negligible computational resources.
During this fine-tuning phase, we utilize both the original in-distribution (ID) data and the newly synthesized out-of-distribution (OOD) data generated through our Multiple Input Mixup (MIM) method.
The objective function used in this fine-tuning step is the following components: 

\begin{align*}
    \mathcal{L} &= CE(M(x_{\text{ID}}), y_{\text{ID}}) + CE(M(x_{\text{OOD}}), \mathcal{P}_{Uniform})
\end{align*}
where $M(x_{\text{ID}})$ and $M(x_{\text{OOD}})$ are the output probability of the model for ID and OOD data, and $y_{\text{ID}}$ represents the true labels for the in-distribution (ID) data. The Cross-Entropy loss for OOD data assumes a uniform distribution as the target~\cite{uniform}.
After utilizing the fine-tuning process, we employ the Maximum Softmax Probability (MSP) technique~\cite{MSP} to perform OOD detection. 
Our empirical results demonstrate that this simple strategy significantly improves OOD detection performance, achieving state-of-the-art results while preserving the original classification accuracy.

\subsection{Data Augmentation}
\vspace{-0.1cm}
With the extensive experiments of our MIM method, We find that employing additional data augmentation techniques after mixup process significantly boosts the OOD detection capabilities of the model.
In our training phase, we apply a comprehensive data augmentation strategy~\cite{imagenet_aug} that includes:
\vspace{-0.1cm}
\begin{itemize}
    \item \textbf{Resizing:} Each image in the dataset is resized to a uniform dimension of $32 \times 32$ pixels. This standardization is crucial for maintaining the consistency of the input data.
    \item \textbf{Color Adjustments:} We apply random color jitter to alter the brightness, contrast, saturation, and hue of the images. This step introduces variability in the color space, helping the model to learn from more diverse images.
    \item \textbf{Affine Transformations:} Random affine transformations are also applied, including rotations within a range of $-90$ to $90$ degrees and translations up to $20\%$ in the image space. These transformations simulate various perspectives and angles for the artificial OOD samples.
\end{itemize}

\begin{table*}[!ht]
\renewcommand\arraystretch{1.0}
\begin{adjustbox}{width=17.8cm,center}
\begin{tabular}{c|c|cccccccccccccc|cc} \toprule
\multirow{4}{*}{\textbf{ID}} & \multirow{4}{*}{\textbf{Methods}} & \multicolumn{14}{c}{\textbf{OOD Datasets}} \\
 &  & \multicolumn{2}{c}{\textbf{Tiny-ImageNet-crop}} & \multicolumn{2}{c}{\textbf{Tiny-ImageNet-resize}} & \multicolumn{2}{c}{\textbf{SVHN}} & \multicolumn{2}{c}{\textbf{LSUN-crop}} & \multicolumn{2}{c}{\textbf{LSUN-resize}} & \multicolumn{2}{c}{\textbf{Textures}} & \multicolumn{2}{c|}{\textbf{Places365}}& \multicolumn{2}{c}{\textbf{Average}} \\
 &  & AUROC & AUPR & AUROC & AUPR & AUROC & AUPR & AUROC & AUPR & AUROC & AUPR & AUROC & AUPR & AUROC & AUPR & AUROC & AUPR\\ 
 &  & $\uparrow$ & $\uparrow$ & $\uparrow$ & $\uparrow$ & $\uparrow$ & $\uparrow$ & $\uparrow$ & $\uparrow$ & $\uparrow$ & $\uparrow$ & $\uparrow$ & $\uparrow$ & $\uparrow$ & $\uparrow$ &$\uparrow$ & $\uparrow$\\ \midrule
\multirow{12}{*}{\textbf{CIFAR-10}} 
 & MSP & 0.9459	&0.9310	&0.8803	&0.8575	&0.9191	&0.9581	&0.9648	&0.9568	&0.9107	&0.8906	&0.8851	&0.7849	&0.8824	&0.9561&0.9126	&0.9050\\
 & ODIN  &0.9580	&0.9518	&0.8970	&0.8891	&0.9162	&0.9594	&0.9748	&0.9730	&0.9277	&0.9213	&0.8834	&0.8069&0.8792	&0.9568	&0.9195	&0.9226\\
 & Mahalanobis   &0.8680	&0.8204	&0.9146	&0.8984	&0.9681	&0.9873	&0.9368	&0.9226	&0.9221&0.8959 &0.9626&\textbf{0.9454}	&0.8016	&0.9231	&0.9105	&0.9133\\
 & Entropy &0.9543	&0.9478	&0.8849	&0.8715	&0.9239	&0.9631	&0.9732	&0.9713	&0.9174	&0.9072	&0.8883	&0.8038&0.8870	&0.9603	&0.9184	&0.9179\\
 & Energy  &0.9797	&0.9775	&0.8890	&0.8841	&0.9107	&0.9603	&0.9905	&0.9902	&0.9382	&0.9345	&0.8534	&0.7847&0.8988	&0.9675	&0.9229	&0.9284\\
 & MaxLogit  &0.9783	&0.9755	&0.8888	&0.8835	&0.9110	&0.9607	&0.9895	&0.9889	&0.9372	&0.9329	&0.8548	&0.7865	&0.8981	&0.9671	&0.9225	&0.9279\\
 & KL-Matching &0.8362	&0.8642	&0.6662	&0.7405	&0.8434	&0.9250	&0.9083	&0.9097	&0.7199	&0.7913	&0.7598	&0.6980	&0.7207	&0.9157	&0.7792	&0.8349\\
 & DICE &0.9811	&0.9794	&0.8760	&0.8726	&0.9126	&0.9606	&0.9915	&0.9913	&0.9320	&0.9290	&0.8458	&0.7757&0.8973	&0.9672	&0.9195	&0.9251\\
 & RMD  &0.9565	&0.9464	&0.8996	&0.8922	&0.9365	&0.9647	&0.9751	&0.9668	&0.9399	&0.9352	&0.9285	&0.8689&\textbf{0.9144}&\textbf{0.9713}&0.9358	&0.9351\\
 & Vim &0.9582	&0.9539	&0.903	&0.8892	&0.9558	&0.9816	&0.9848	&0.9845	&0.9286	&0.9167	&0.9378&0.9089&0.8559	&0.9481	&0.9320	&0.9404\\
  \cline{2-18}
 & Ours   &\textbf{0.9873}&\textbf{0.9867}&\textbf{0.9649}&\textbf{0.9589}&\textbf{0.9726}&\textbf{0.9881}&\textbf{0.9952}&\textbf{0.9950}&\textbf{0.9752}&\textbf{0.9698}&\textbf{0.9627}	&0.9419&0.9105	&0.9695	&\textbf{0.9669}&\textbf{0.9728}\\ 
\hline
 \multirow{12}{*}{\textbf{CIFAR-100}} 
 & MSP  &0.8632	&0.8480	&0.7464	&0.7091	&0.7137 &0.8437&0.8558 &0.8435&0.7537&0.7116	&0.7356	&0.5750&0.7391	&0.8944	&0.7725	&0.7750\\
 & ODIN  &0.8736	&0.8623	&0.7790	&0.7624	&0.6469	&0.7865	&0.8569	&0.8468	&0.7864	&0.7651	&0.7262	&0.5728&0.7309	&0.8894	&0.7714	&0.7836\\
 & Mahalanobis &0.5595	&0.4922	&0.9330	&0.9321	&0.8568	&0.9309	&0.5203	&0.4674	&0.9443	&0.9404	&0.8993	&0.8529	&0.6393	&0.8425	&0.7646	&0.7798\\
 & Entropy  &0.9191	&0.9206	&0.6394	&0.6394	&0.5265	&0.7236	&0.8961	&0.8989	&0.6603	&0.5927	&0.5466	&0.3955&0.6489	&0.8600	&0.6910	&0.7187\\
 & Energy &0.9592 &0.9581 &0.5618 &0.5190&0.4828 &0.6966 &\textbf{0.9691} &\textbf{0.9712} &0.5649 &0.5137	&0.5353	&0.3970 &0.6349 &0.8432 &0.6726 &0.6998\\
 & MaxLogit  &0.9467	&0.9401	&0.7769	&0.7421	&0.7395	&0.8536	&0.9514	&0.9468	&0.7932	&0.7519	&0.7637	&0.6164&0.7594	&0.9050	&0.8187	&0.8223\\
 & KL-Matching  &0.8209	&0.8180	&0.7635	&0.7662	&0.7014	&0.8676	&0.7772	&0.8041	&0.7814&0.7876	&0.7189	&0.6310&0.6606	&0.8807	&0.7463	&0.7936\\
 & DICE  &0.9566	&0.9542	&0.7821	&0.7495	&0.7402	&0.8516	&0.9630	&0.9622	&0.8022	&0.7655	&0.7649	&0.6213&0.7630	&0.9074	&0.8246	&0.8302\\
 & RMD  &0.8737	&0.8223	&0.8975	&0.8954	&0.8021	&0.9098	&0.8543	&0.8138&0.9220	&0.9210	&0.8671	&0.8074&\textbf{0.7859}&\textbf{0.9196}&0.8575	&0.8699\\
 & Vim &0.9069&0.8626&\textbf{0.7555}&\textbf{0.6573}&\textbf{0.9474}&\textbf{0.9482}&0.7996	&0.7154	&\textbf{0.9455}&\textbf{0.9476}&\textbf{0.9254}&\textbf{0.9664}&0.7011	&0.8768	&0.8545	&.8535\\
  \cline{2-18}
 & Ours &\textbf{0.9667}&\textbf{0.9617}&0.9171	&0.8960	&0.7936	&0.8810	&0.9673	&0.9639	&0.9149	&0.8878	&0.8603	&0.7850&0.7409	&0.8937	&\textbf{0.8801}&\textbf{0.8956}\\ 
 \bottomrule
\end{tabular}
\end{adjustbox}
\caption{Comparison with state-of-the-art methods using a WideResNet-40-2 classifier. All experiments are conducted by the OOD detection benchmark framework~\cite{kirchheim2022pytorch}. The symbol $\uparrow$ indicates larger values are better.}
\label{tab:cifar10_table}
\end{table*}
\vspace{-0.1cm}
\begin{figure*}[h]
    \centering \centerline{\includegraphics[width=0.85\textwidth]{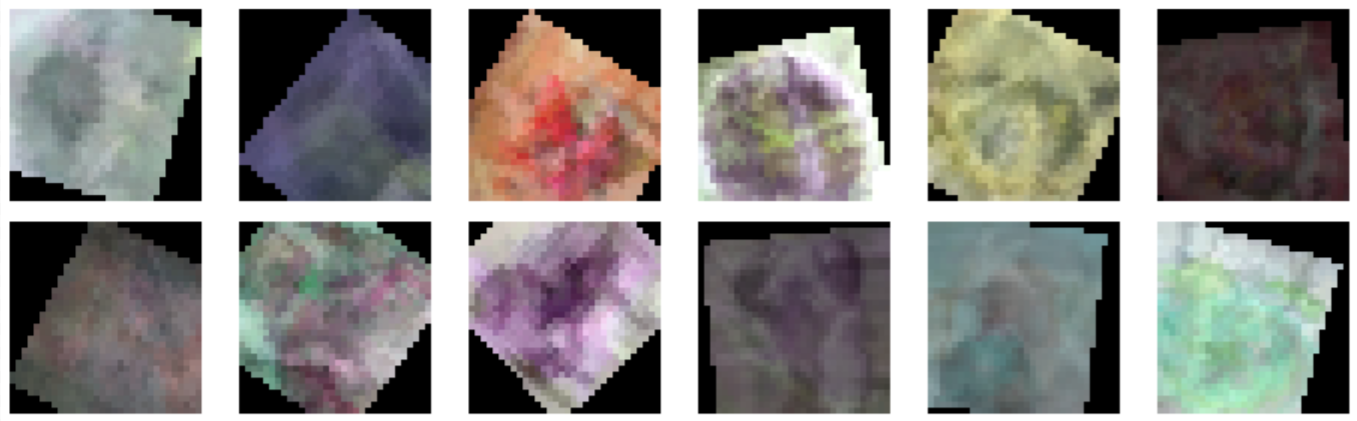}}
    \caption{The image examples generated from our proposed method. The generated samples consist of marginal features, which are greatly useful for training the OOD detection network.} 
    \label{fig:mixup_augmentation}
\end{figure*}

\section{Experiments}
\vspace{-0.1cm}
\subsection{Datasets}
\vspace{-0.1cm}
For experiments, we select the broadly used in-distribution datasets, CIFAR-10 and CIFAR-100~\cite{cifar}.
For OOD datasets, we have adopted 7 OOD datasets for the CIFAR10 and CIFAR100 ID setup.
\vspace{-0.1cm}
\begin{itemize}
    \item \textbf{Textures} This dataset contains 5,640 natural texture images categorized into 47 groups. Each category consists of 120 images, ranging in their size from 300$\times$300 to 640$\times$640 pixels~\cite{textures}.
    \item \textbf{LSUN-crop} As a component of the Large-scale Scene Understanding (LSUN) project, this dataset consists of various cropped scene images, labeled using a hybrid of automated and manual processes~\cite{LSUN}.
    \item \textbf{LSUN-resize} Another component of the LSUN project, this dataset aims to enhance visual recognition with images of different sizes~\cite{LSUN}.
    \item \textbf{Tiny-ImageNet-crop}  A modified version of Tiny-ImageNet, this dataset consists of images cropped from the original ImageNet dataset~\cite{TinyImageNet}.
    \item \textbf{Tiny-ImageNet-resize} Another variation of Tiny-ImageNet, this dataset includes images resized to various dimensions, providing a resource to enhance the scale adaptability of models~\cite{TinyImageNet}.
    \item \textbf{SVHN} Sourced from Google Street View house numbers, this dataset contains small and cropped digit images~\cite{SVHN}.
    \item \textbf{Places365} Designed for scene recognition, this dataset comprises various scene images across 434 different scene categories~\cite{Places365}. 
\end{itemize}
\vspace{-0.1cm}
\subsection{Evaluation Metrics}
\vspace{-0.13cm}
In our evaluation, we employ two essential metrics to assess the performance of our model:
\vspace{-0.1cm}
\begin{itemize}
    \item \textbf{AUROC (Area Under the Receiver Operating Characteristic):} This metric~\cite{AUROC} quantifies a model's ability to distinguish between ID and OOD samples based on the ROC curve, providing a comprehensive measure of OOD detection performance.
    \item \textbf{AUPR (Area Under the Precision-Recall Curve):} AUPR~\cite{AUPR} evaluates the precision-recall trade-off in OOD detection. This measure provides insights into how effectively our model balances precision and recall when identifying OOD data.
\end{itemize}
\vspace{-0.1cm}
\subsection{Training Details}
\vspace{-0.1cm}
In our experimental setup, we utilize the WideResNet-40-2 architecture~\cite{WideResnet} for training our models, with CIFAR-10 and CIFAR-100 serving as our in-distribution (ID) datasets. As a following step, we evaluate out-of-distribution (OOD) detection performance using the seven diverse datasets we previously referred to, comparing our models' effectiveness against various SOTA counterparts. Throughout our OOD detection and evaluation experiments, we employ a specialized OOD detection framework, ensuring a comprehensive and robust evaluation against diverse datasets~\cite{kirchheim2022pytorch}. 

Specifically for our model, we implement an additional single fine-tuning epoch using our method. During this epoch, we apply stochastic gradient descent (SGD) optimization~\cite{SGD} with a learning rate of 0.0001 and a momentum of 0.9. We also experiment with varying the number of images for mixup and find that using a number ranging from 5 to 10 shows improved overall performance.

\begin{table*}[!ht]
\renewcommand\arraystretch{1.0}
\begin{adjustbox}{width=17.8cm,center}
\begin{tabular}{c|c|ccccccc|c} \toprule
\multirow{3}{*}{\textbf{ID}} & \multirow{3}{*}{\textbf{Methods}} & \multicolumn{7}{c}{\textbf{OOD Datasets}} \\
 &  & \multicolumn{1}{c}{\textbf{Tiny-ImageNet-crop}} & \multicolumn{1}{c}{\textbf{Tiny-ImageNet-resize}} & \multicolumn{1}{c}{\textbf{SVHN}} & \multicolumn{1}{c}{\textbf{LSUN-crop}} & \multicolumn{1}{c}{\textbf{LSUN-resize}} & \multicolumn{1}{c}{\textbf{Textures}} & \multicolumn{1}{c|}{\textbf{Places365}}& \multicolumn{1}{c}{\textbf{Average}} \\
 &  & $\uparrow$ & $\uparrow$ & $\uparrow$ & $\uparrow$ & $\uparrow$ & $\uparrow$ & $\uparrow$ & $\uparrow$\\ \midrule
\multirow{2}{*}{\textbf{CIFAR-10}} 
 & OE  &0.9880	&0.9731	&0.9841	&0.9969	&0.9894	&0.9775	&0.9637	&0.9818\\
\cline{2-10}
 & Ours &0.9889	&0.9696	&0.9750	&0.9957	&0.9789	&0.9643	&0.9141	&0.9695\\ 
 \hline
 \multirow{2}{*}{\textbf{CIFAR-100}} 
 & OE   &0.8925	&0.5356	&0.6879	&0.9341	&0.6094	&0.671	&0.698	&0.7184\\
\cline{2-10}
 & Ours &0.9667	&0.9171	&0.7936	&0.9673	&0.9149	&0.8603	&0.7410	&0.8801\\ 
 \bottomrule
\end{tabular}
\end{adjustbox}
\caption{OOD detection performance (AUROC) comparison results of the Outlier Exposure (OE) and \textbf{Ours}.}
\label{tab:OE_comparison_table}
\end{table*}
\vspace{-0.1cm}

\subsection{Experimental Results}
\textbf{Overall OOD Performance}\indent Interestingly, we have observed that our proposed method significantly outperforms existing SOTA methods in out-of-distribution (OOD) detection despite its extremely simple approach. Our model's superior performance has been reported in Table~\ref{tab:cifar10_table}, which presents a comparative analysis of average AUROC (Area Under the Receiver Operating Characteristic) and AUPR (Area Under the Precision-Recall Curve) metrics across various datasets. We note that this result has not been cherry-picked; the proposed method consistently demonstrates higher OOD performance in various scenarios.

\smallskip
\noindent\textbf{Impact on ID Accuracy}\indent Additionally, a notable advantage of our method is its minimal impact on the original ID test accuracy. By additionally training a single epoch fine-tuning with existing ID data, our approach results in negligible changes in ID test accuracy. On CIFAR-10, the test accuracy shifts slightly from 94.84\% to 94.47\%, and on CIFAR-100, it alters from 75.95\% to 74.41\%. These minor variations confirm that our method does not significantly affect the model's ID test accuracy. In contrast, the AUROC performance of OOD detection shows a remarkable increase, jumping from an average of 0.912 to 0.966 on CIFAR-10 and 0.772 to 0.88 on CIFAR-100. These significant improvements establish our method as the highest performance in overall scores for both CIFAR-10 and CIFAR-100 datasets, demonstrating its superior capability in OOD detection.

\smallskip
\noindent\textbf{Comparison with Outlier Exposure}\indent Furthermore, when compared to the Outlier Exposure (OE) methodology, our approach has demonstrated robust OOD performance. As shown in Table~\ref{tab:OE_comparison_table}, our method exceeds the AUROC of OE by 0.16 in the CIFAR-100 experiment. We emphasize the efficiency of our approach: our method achieves competitive performance with OE without the need to collect thousands of additional data points. Our method simply utilizes processed and generated OOD data, proving the effectiveness of our lightweight and practical approach to OOD detection.

\vspace{-0.1cm}
\section{Conclusion}
In this study, we introduce the Multiple Input Mixup (MIM) method, a straightforward yet effective approach to enhance OOD detection performance for deep learning models. MIM requires a single additional training epoch on a pre-trained classifier and demonstrates superior OOD detection capabilities, as shown in our CIFAR-10 and CIFAR-100 experiments. Notably, while maintaining original ID test accuracy, MIM outperforms SOTA Outlier Exposure (OE) methods in the CIFAR-100 benchmark. This balance of simplicity, efficiency, and effectiveness establishes MIM as a promising tool for reliable AI model deployment across various domains. Our work highlights the potential of simple, yet powerful methods in AI safety and reliability, encouraging further research in this crucial OOD detection research area.
\vspace{-0.2cm}
\section{Acknowledgements}
This research was supported by Brian Impact, a non-profit organization dedicated to advancing science and technology

\bibliography{aaai24}

\begin{thebibliography}{31}
\providecommand{\natexlab}[1]{#1}

\bibitem[{Bottou(2012)}]{SGD}
Bottou, L. 2012.
\newblock Stochastic gradient descent tricks.
\newblock In \emph{Neural Networks: Tricks of the Trade: Second Edition}, 421--436. Springer.

\bibitem[{Chan, Rottmann, and Gottschalk(2021)}]{entropy}
Chan, R.; Rottmann, M.; and Gottschalk, H. 2021.
\newblock Entropy maximization and meta classification for out-of-distribution detection in semantic segmentation.
\newblock In \emph{Proceedings of the ieee/cvf international conference on computer vision}, 5128--5137.

\bibitem[{Cimpoi et~al.(2014)Cimpoi, Maji, Kokkinos, Mohamed, and Vedaldi}]{textures}
Cimpoi, M.; Maji, S.; Kokkinos, I.; Mohamed, S.; and Vedaldi, A. 2014.
\newblock Describing textures in the wild.
\newblock In \emph{Proceedings of the IEEE conference on computer vision and pattern recognition}, 3606--3613.

\bibitem[{Davis and Goadrich(2006)}]{AUPR}
Davis, J.; and Goadrich, M. 2006.
\newblock The relationship between Precision-Recall and ROC curves.
\newblock In \emph{Proceedings of the 23rd international conference on Machine learning}, 233--240.

\bibitem[{Hanley and McNeil(1982)}]{AUROC}
Hanley, J.~A.; and McNeil, B.~J. 1982.
\newblock The meaning and use of the area under a receiver operating characteristic (ROC) curve.
\newblock \emph{Radiology}, 143(1): 29--36.

\bibitem[{Hendrycks et~al.(2019{\natexlab{a}})Hendrycks, Basart, Mazeika, Zou, Kwon, Mostajabi, Steinhardt, and Song}]{KLmatchingLogits}
Hendrycks, D.; Basart, S.; Mazeika, M.; Zou, A.; Kwon, J.; Mostajabi, M.; Steinhardt, J.; and Song, D. 2019{\natexlab{a}}.
\newblock Scaling out-of-distribution detection for real-world settings.
\newblock \emph{arXiv preprint arXiv:1911.11132}.

\bibitem[{Hendrycks et~al.(2019{\natexlab{b}})Hendrycks, Basart, Mazeika, Zou, Kwon, Mostajabi, Steinhardt, and Song}]{MaxLigit}
Hendrycks, D.; Basart, S.; Mazeika, M.; Zou, A.; Kwon, J.; Mostajabi, M.; Steinhardt, J.; and Song, D. 2019{\natexlab{b}}.
\newblock Scaling out-of-distribution detection for real-world settings.
\newblock \emph{arXiv preprint arXiv:1911.11132}.

\bibitem[{Hendrycks and Gimpel(2016{\natexlab{a}})}]{MSP}
Hendrycks, D.; and Gimpel, K. 2016{\natexlab{a}}.
\newblock A baseline for detecting misclassified and out-of-distribution examples in neural networks.
\newblock \emph{arXiv preprint arXiv:1610.02136}.

\bibitem[{Hendrycks and Gimpel(2016{\natexlab{b}})}]{uniform}
Hendrycks, D.; and Gimpel, K. 2016{\natexlab{b}}.
\newblock A baseline for detecting misclassified and out-of-distribution examples in neural networks.
\newblock \emph{arXiv preprint arXiv:1610.02136}.

\bibitem[{Hendrycks, Mazeika, and Dietterich(2018)}]{OE}
Hendrycks, D.; Mazeika, M.; and Dietterich, T. 2018.
\newblock Deep anomaly detection with outlier exposure.
\newblock \emph{arXiv preprint arXiv:1812.04606}.

\bibitem[{Kim et~al.(2023)Kim, Kong, Na, and Jung}]{kirby}
Kim, J.; Kong, S.~T.; Na, D.; and Jung, K.-H. 2023.
\newblock Key feature replacement of in-distribution samples for out-of-distribution detection.
\newblock In \emph{Proceedings of the AAAI Conference on Artificial Intelligence}, volume~37, 8246--8254.

\bibitem[{Kirchheim, Filax, and Ortmeier(2022)}]{kirchheim2022pytorch}
Kirchheim, K.; Filax, M.; and Ortmeier, F. 2022.
\newblock PyTorch-OOD: A Library for Out-of-Distribution Detection Based on PyTorch.
\newblock In \emph{Proceedings of the IEEE/CVF Conference on Computer Vision and Pattern Recognition (CVPR) Workshops}, 4351--4360.

\bibitem[{Krizhevsky, Hinton et~al.(2009)}]{cifar}
Krizhevsky, A.; Hinton, G.; et~al. 2009.
\newblock Learning multiple layers of features from tiny images.

\bibitem[{Lee et~al.(2017)Lee, Lee, Lee, and Shin}]{gan}
Lee, K.; Lee, H.; Lee, K.; and Shin, J. 2017.
\newblock Training confidence-calibrated classifiers for detecting out-of-distribution samples.
\newblock \emph{arXiv preprint arXiv:1711.09325}.

\bibitem[{Lee et~al.(2018)Lee, Lee, Lee, and Shin}]{mahalanobis}
Lee, K.; Lee, K.; Lee, H.; and Shin, J. 2018.
\newblock A simple unified framework for detecting out-of-distribution samples and adversarial attacks.
\newblock \emph{Advances in neural information processing systems}, 31.

\bibitem[{Liang, Li, and Srikant(2017{\natexlab{a}})}]{liang2017enhancing}
Liang, S.; Li, Y.; and Srikant, R. 2017{\natexlab{a}}.
\newblock Enhancing the reliability of out-of-distribution image detection in neural networks.
\newblock \emph{arXiv preprint arXiv:1706.02690}.

\bibitem[{Liang, Li, and Srikant(2017{\natexlab{b}})}]{odin}
Liang, S.; Li, Y.; and Srikant, R. 2017{\natexlab{b}}.
\newblock Enhancing the reliability of out-of-distribution image detection in neural networks.
\newblock \emph{arXiv preprint arXiv:1706.02690}.

\bibitem[{Liu et~al.(2020)Liu, Wang, Owens, and Li}]{energy}
Liu, W.; Wang, X.; Owens, J.; and Li, Y. 2020.
\newblock Energy-based out-of-distribution detection.
\newblock \emph{Advances in neural information processing systems}, 33: 21464--21475.

\bibitem[{L{\'o}pez-Cifuentes et~al.(2020)L{\'o}pez-Cifuentes, Escudero-Vinolo, Besc{\'o}s, and Garc{\'\i}a-Mart{\'\i}n}]{Places365}
L{\'o}pez-Cifuentes, A.; Escudero-Vinolo, M.; Besc{\'o}s, J.; and Garc{\'\i}a-Mart{\'\i}n, {\'A}. 2020.
\newblock Semantic-aware scene recognition.
\newblock \emph{Pattern Recognition}, 102: 107256.

\bibitem[{Netzer et~al.(2011)Netzer, Wang, Coates, Bissacco, Wu, and Ng}]{SVHN}
Netzer, Y.; Wang, T.; Coates, A.; Bissacco, A.; Wu, B.; and Ng, A.~Y. 2011.
\newblock Reading digits in natural images with unsupervised feature learning.

\bibitem[{Ren et~al.(2021)Ren, Fort, Liu, Roy, Padhy, and Lakshminarayanan}]{RMD}
Ren, J.; Fort, S.; Liu, J.; Roy, A.~G.; Padhy, S.; and Lakshminarayanan, B. 2021.
\newblock A simple fix to mahalanobis distance for improving near-ood detection.
\newblock \emph{arXiv preprint arXiv:2106.09022}.

\bibitem[{Russakovsky et~al.(2015)Russakovsky, Deng, Su, Krause, Satheesh, Ma, Huang, Karpathy, Khosla, Bernstein et~al.}]{imagenet_aug}
Russakovsky, O.; Deng, J.; Su, H.; Krause, J.; Satheesh, S.; Ma, S.; Huang, Z.; Karpathy, A.; Khosla, A.; Bernstein, M.; et~al. 2015.
\newblock Imagenet large scale visual recognition challenge.
\newblock \emph{International journal of computer vision}, 115: 211--252.

\bibitem[{Sun, Guo, and Li(2021)}]{TinyImageNet}
Sun, Y.; Guo, C.; and Li, Y. 2021.
\newblock React: Out-of-distribution detection with rectified activations.
\newblock \emph{Advances in Neural Information Processing Systems}, 34: 144--157.

\bibitem[{Sun and Li(2022)}]{DICE}
Sun, Y.; and Li, Y. 2022.
\newblock DICE: Leveraging Sparsification for Out-of-Distribution Detection.
\newblock In \emph{European Conference on Computer Vision}.

\bibitem[{Verma et~al.(2019)Verma, Lamb, Beckham, Najafi, Mitliagkas, Lopez-Paz, and Bengio}]{manifoldmixup}
Verma, V.; Lamb, A.; Beckham, C.; Najafi, A.; Mitliagkas, I.; Lopez-Paz, D.; and Bengio, Y. 2019.
\newblock Manifold mixup: Better representations by interpolating hidden states.
\newblock In \emph{International conference on machine learning}, 6438--6447. PMLR.

\bibitem[{Wang et~al.(2022)Wang, Li, Feng, and Zhang}]{vim}
Wang, H.; Li, Z.; Feng, L.; and Zhang, W. 2022.
\newblock Vim: Out-of-distribution with virtual-logit matching.
\newblock In \emph{Proceedings of the IEEE/CVF conference on computer vision and pattern recognition}, 4921--4930.

\bibitem[{Yosinski et~al.(2014)Yosinski, Clune, Bengio, and Lipson}]{finetuning}
Yosinski, J.; Clune, J.; Bengio, Y.; and Lipson, H. 2014.
\newblock How transferable are features in deep neural networks?
\newblock \emph{Advances in neural information processing systems}, 27.

\bibitem[{Yu et~al.(2015)Yu, Seff, Zhang, Song, Funkhouser, and Xiao}]{LSUN}
Yu, F.; Seff, A.; Zhang, Y.; Song, S.; Funkhouser, T.; and Xiao, J. 2015.
\newblock Lsun: Construction of a large-scale image dataset using deep learning with humans in the loop.
\newblock \emph{arXiv preprint arXiv:1506.03365}.

\bibitem[{Yun et~al.(2019)Yun, Han, Oh, Chun, Choe, and Yoo}]{cutmix}
Yun, S.; Han, D.; Oh, S.~J.; Chun, S.; Choe, J.; and Yoo, Y. 2019.
\newblock Cutmix: Regularization strategy to train strong classifiers with localizable features.
\newblock In \emph{Proceedings of the IEEE/CVF international conference on computer vision}, 6023--6032.

\bibitem[{Zagoruyko and Komodakis(2016)}]{WideResnet}
Zagoruyko, S.; and Komodakis, N. 2016.
\newblock Wide residual networks.
\newblock \emph{arXiv preprint arXiv:1605.07146}.

\bibitem[{Zhang et~al.(2017)Zhang, Cisse, Dauphin, and Lopez-Paz}]{mixup}
Zhang, H.; Cisse, M.; Dauphin, Y.~N.; and Lopez-Paz, D. 2017.
\newblock mixup: Beyond empirical risk minimization.
\newblock \emph{arXiv preprint arXiv:1710.09412}.

\end{thebibliography}

\end{document}